# MSA-technique for stiffness modeling of manipulators with complex and hybrid structures

A. Klimchik*, A. Pashkevich*,**, D. Chablat***

*Innopolis University, Universitetskaya 1, 420500 Innopolis, The Republic of Tatarstan, Russia
(e-mail: a.klimchik@innopolis.ru)
** IMT Atlantique, 4 rue Alfred-Kastler, Nantes 44307, Le Laboratoire des Sciences du Numérique de Nantes (LS2N)
(e-mail: anatol.pashkevich@imt-atlantique.fr)
*** CNRS, Nantes, France, Le Laboratoire des Sciences du Numérique de Nantes (LS2N)
(e-mail: damien.chablat@cnrs.fr)

**Abstract:** The paper presents a systematic approach for stiffness modeling of manipulators with complex and hybrid structures using matrix structural analysis. In contrast to previous results, it is suitable for mixed architectures containing closed-loops, flexible links, rigid connections, passive and elastic joints with external loadings and preloadings. The proposed approach produces the Cartesian stiffness matrices in a semi-analytical manner. It presents the manipulator stiffness model as a set of conventional equations describing the link elasticities that are supplemented by a set of constraints describing connections between links. Its allows user straightforward aggregation of stiffness model equations avoiding traditional column/row merging procedures in the extended stiffness matrix. Advantages of this approach are illustrated by stiffness analysis of NaVaRo manipulator.

## 1. INTRODUCTION

In many modern robotic applications, manipulators are subject to essential external loadings that affect the positioning accuracy and provoke non-negligible compliance errors (Alici and Shirinzadeh, 2005, Klimchik et al. , 2017, Slavkovic et al. , 2013). For this reason, manipulator stiffness analysis becomes one of the most important issues in the design of robot mechanics and control algorithms. It allows the designer to achieve required balance between the dynamics and accuracy. However, to make the stiffness analysis efficient, it should rely on a simple and computationally reasonable method that is able to deal with complex architectures including closed-loops, flexible links, and rigid connections, passive and elastic joints that are common for parallel robots.

At present, there exist three main techniques in this area: Finite Element Analysis (FEA), Matrix Structural Analysis (MSA) and Virtual Joint Modeling (VJM) (Gosselin and Zhang, 2002, Klimchik et al. , 2014, Pashkevich et al. , 2011, Quennouelle and Gosselin, 2008). The most accurate of them is the FEA (Wang et al. , 2006, Yan et al. , 2016), which allows modeling links and joints with their true dimension and shape. However, it is usually applied at the final design stage because of the high computational expenses required for high order matrix inversion. The MSA is considered as a compromise technique, which incorporates the main ideas of the FEA, but operates with rather large elements such as flexible links connected by the actuated and passive joints in the overall manipulator structure (Przemieniecki, 1985). This obviously leads to the reduction of the computational expenses that are quite acceptable for robotics, but it requires some non-trivial actions for including of passive and elastic joints in the related mathematical model. From the other side, the MSA is very convenient for a description of complex structures with numerous closed-loops and cross-linkage.

Some reviews of existing works on manipulator stiffness analysis can be found in (Gonçalves et al. , 2016, Klimchik, Chablat, 2014, Klimchik et al. , 2015, Liu et al. , 2017, Pinto et al. , 2009, Taghvaeipour et al. , 2012, Yeo et al. , 2013), where the authors addressed different aspects of the FEA, MSA and VJM techniques and particularities of their application in robotics. This works cover results starting from the early works of Salisbury and Gosselin (Gosselin, 1990, Salisbury, 1980) till nowadays, excluding last few years. Among recent contributions devoted to the MSA it worth mentioning the work of Cammarata (Cammarata, 2016), who introduced the notion of the Condensed Stiffness Matrix that merges together flexibilities of the link and adjacent joints. Further, the obtained stiffness matrices of a two-node super-elements are used in a traditional for MSA way, which includes manual work with merging lines and columns of the global stiffness matrix. Besides, despite the apparent simplicity, this technique does not allow direct computing of the Cartesian Stiffness Matrix, which is the main outcome characteristic of the manipulator stiffness analysis. In addition, the loading mode and buckling analysis cannot be treated in a simple way. A useful extension of the MSA for the case of links and joints with non-linear stiffness properties was proposed in (Detert and Corves, 2017). In particular, a passive revolute joint with ball bearings was presented as an element with rank-deficient force-dependent stiffness matrix, whose parameters were estimated experimentally. The relevant computational procedure included several iterations of conventional MSA linear model. This technique was applied to PARAGRIP handling system and validated by measurements of the end-effector deflection under vertical load. There are also several works the deal with the MSA application to the stiffness analysis of particular parallel and serial manipulators (Azulay et al. , 2014, Koseki et al. , 2002, Li et al. , 2011, Marie et al. , 2013, Shi et al. , 2017).

To our knowledge, the most essential contribution to the robot-oriented modification of the MSA was done by Deblase et. all (Deblaise et al. , 2006). The authors proposed a general technique to take into account passive joints and rigid connections, which are common for parallel robots, without changing $12 \times 12$ stiffness matrices of structural elements. Nevertheless, some manual procedures of merging matrix components were not avoided, as well as preloadings and elastic connections were not treated. For this reason, this paper focuses on some enhancement and generalization of the MSA technique for robotic applications providing a compromise between complexity of the stiffness model generation and complexity of subsequent computations.

## 2. MSA MODELS OF MANIPULATOR LINKS AND PLATFORMS

The proposed technique is based on the separate description of the stiffness model components and connections between them, which further aggregated algorithmically. It allows avoiding manual merging of the stiffness matrix rows/columns, replacing this operation by numerical matrix computations. Let us present first the stiffness model components considered in this paper.

### 2.1. Modeling of flexible link

If the link flexibility is non-negligible, the 2-node "free-free" stiffness model should be used, which can be obtained either from the approximation of the link by a beam or using the CAD environment with integrated FEA toolbox (Klimchik et al. , 2013). In both cases, the link is described by the linear matrix equation

$$\begin{bmatrix} \mathbf{W}_i \\ \mathbf{W}_j \end{bmatrix} = \begin{bmatrix} \mathbf{K}_{11}^{(ij)} & \mathbf{K}_{12}^{(ij)} \\ \mathbf{K}_{21}^{(ij)} & \mathbf{K}_{22}^{(ij)} \end{bmatrix}_{12 \times 12} \cdot \begin{bmatrix} \Delta \mathbf{t}_i \\ \Delta \mathbf{t}_j \end{bmatrix} \quad (1)$$

where $\Delta \mathbf{t}_i$, $\Delta \mathbf{t}_j$ are the deflections at the link ends, $\mathbf{W}_i$, $\mathbf{W}_j$ are the link end wrenches, $i$ and $j$ are the node indices corresponding to the link ends, and $\mathbf{K}_{11}^{(ij)}$, $\mathbf{K}_{12}^{(ij)}$, $\mathbf{K}_{21}^{(ij)}$, $\mathbf{K}_{22}^{(ij)}$ are $6 \times 6$ stiffness matrices. The linear relation (1) should be represented in a global coordinate system, which is achieve by simple rotation of the local stiffness matrices.

### 2.2. Modeling of rigid link

If the link flexibility is negligible, the above stiffness model should be replaced by a "rigidity constraint". Applying to this link the rigid body kinematic equation, one can get the following relations between the nodal displacements $\Delta \mathbf{t}_i = [\Delta \mathbf{p}_i; \Delta \boldsymbol{\varphi}_i]$, $\Delta \mathbf{t}_j = [\Delta \mathbf{p}_j; \Delta \boldsymbol{\varphi}_j]$

$$\begin{bmatrix} \mathbf{I}_{3\times3} & [\mathbf{d}^{(ij)}\times]^T & -\mathbf{I}_{3\times3} & \mathbf{0}_{3\times3} \\ \mathbf{0}_{3\times3} & \mathbf{I}_{3\times3} & \mathbf{0}_{3\times3} & -\mathbf{I}_{3\times3} \end{bmatrix} \cdot \begin{bmatrix} \Delta \mathbf{t}_i \\ \Delta \mathbf{t}_j \end{bmatrix} = \mathbf{0}_{6\times1} \quad (2)$$

where $[\mathbf{d}^{(ij)}\times]$ denotes the $3 \times 3$ skew-symmetric matrix derived from the vector $\mathbf{d}^{(ij)} = [d_x^{(ij)}, d_y^{(ij)}, d_z^{(ij)}]^T$ describing the link geometry from the $i^{th}$ to $j^{th}$ node. Further, after definition $6 \times 6$ block matrix

$$\mathbf{D}^{(ij)} = \begin{bmatrix} \mathbf{I}_{3\times3} & [\mathbf{d}^{(ij)}\times]^T \\ \mathbf{0}_{3\times3} & \mathbf{I}_{3\times3} \end{bmatrix}_{6\times6} \quad (3)$$

the rigid-link constraint can be presented in the form.

$$\begin{bmatrix} \mathbf{D}^{(ij)} & -\mathbf{I}_{6\times6} \end{bmatrix} \cdot \begin{bmatrix} \Delta \mathbf{t}_i \\ \Delta \mathbf{t}_j \end{bmatrix} = \mathbf{0}_{6\times1} \quad (4)$$

that is convenient for aggregation of the model components.

The second group of equations describing the force equilibrium can be derived using rigid body static equations, which yields the following relations

$$\mathbf{M}_j + \mathbf{M}_i + \mathbf{d}^{(ij)} \times \mathbf{F}_j = 0; \qquad \mathbf{F}_j + \mathbf{F}_i = 0 \quad (5)$$

where $\mathbf{F}_i$, $\mathbf{F}_j$ and $\mathbf{M}_i$, $\mathbf{M}_j$ denote the forces and torques applied at the nodes $i$ and $j$ respectively. In a matrix form, they can be presented as follows

$$\mathbf{W}_i + \begin{bmatrix} \mathbf{I} & 0 \\ [\mathbf{d}^{(ij)}\times] & \mathbf{I} \end{bmatrix} \mathbf{W}_j = 0 \quad (6)$$

Similarly, using the above-introduced definition for $\mathbf{D}^{(ij)}$, this equation can be rewritten in a more compact form

$$\begin{bmatrix} \mathbf{I}_{6\times6} & \mathbf{D}^{(ij)T} \end{bmatrix} \cdot \begin{bmatrix} \mathbf{W}_i \\ \mathbf{W}_j \end{bmatrix} = \mathbf{0}_{6\times1} \quad (7)$$

Hence, the rigid link model includes 12 scalar equations describing the displacement and force equilibriums, which are written with respect to 24 scalar variables contained in six-dimensional vectors $\Delta \mathbf{t}_i$, $\Delta \mathbf{t}_j$ and $\mathbf{W}_i$, $\mathbf{W}_j$, similar to the flexible link case (the rank deficiency of the link model matrix is also equal to 12).

### 2.3. Modeling of rigid mobile platform

If the platform can be treated as a rigid body, it can be included in the global stiffness model by means of several virtual rigid links connecting the nodes $i, j, k, ...$ of the manipulator leg clamping and the virtual rigid node $e$ corresponding to the manipulator end-effector reference point. All of these links produce constraints similar to (4) that may be aggregated in a common matrix equation

$$\begin{bmatrix} \mathbf{D}^{(ie)} & \mathbf{0}_{6\times6} & \mathbf{0}_{6\times6} & -\mathbf{I}_{6\times6} \\ \mathbf{0}_{6\times6} & \mathbf{D}^{(je)} & \mathbf{0}_{6\times6} & -\mathbf{I}_{6\times6} \\ \mathbf{0}_{6\times6} & \mathbf{0}_{6\times6} & \mathbf{D}^{(ke)} & -\mathbf{I}_{6\times6} \end{bmatrix} \cdot \begin{bmatrix} \Delta \mathbf{t}_i \\ \Delta \mathbf{t}_j \\ \Delta \mathbf{t}_k \\ \Delta \mathbf{t}_e \end{bmatrix} = \mathbf{0}_{18\times1} \quad (8)$$

where $\mathbf{D}^{(ie)}$, $\mathbf{D}^{(je)}$, $\mathbf{D}^{(ke)}$ describe the virtual links geometry in the same way as in (3).

However, the force equilibrium produces here a single equation only

$$\mathbf{D}^{(ei)T} \mathbf{W}_i + \mathbf{D}^{(ej)T} \mathbf{W}_j + \mathbf{D}^{(ek)T} \mathbf{W}_k + \mathbf{W}_e = \mathbf{0}_{6\times1} \quad (9)$$

which is an extended form of (7). In a matrix form, it can be presented as

$$\begin{bmatrix} \mathbf{D}^{(ei)T} & \mathbf{D}^{(ej)T} & \mathbf{D}^{(ek)T} & \mathbf{I}_{6\times6} \end{bmatrix} \cdot col(\mathbf{W}_i, \mathbf{W}_j, \mathbf{W}_k, \mathbf{W}_e) = \mathbf{0}_{6\times1} \quad (10)$$

Here, the number of the leg clamping points is assumed to be equal to three, but it can be easily generalized. Hence, for this example, the model includes 24 scalar equations and 48 scalar variables. The rank deficiency of the model matrix is equal to 24, which agrees with the rigid body mechanics.

## 2.4. Modeling of non-rigid mobile platform

If the platform cannot be treated as a rigid body, it should be approximated by several virtual flexible links connecting the nodes $i, j, k, ...$ of the manipulator leg clamping and the virtual rigid node $e$. These links are described by $12 \times 12$ stiffness matrices (1) that are aggregated in a matrix equation

$$\begin{bmatrix} \mathbf{W}_i \\ \mathbf{W}_j \\ \mathbf{W}_k \\ \mathbf{W}_e \end{bmatrix} = \begin{bmatrix} \mathbf{K}_{11}^{(ie)} & \mathbf{0}_{6\times 6} & \mathbf{0}_{6\times 6} & \mathbf{K}_{12}^{(ie)} \\ \mathbf{0}_{6\times 6} & \mathbf{K}_{11}^{(je)} & \mathbf{0}_{6\times 6} & \mathbf{K}_{12}^{(je)} \\ \mathbf{0}_{6\times 6} & \mathbf{0}_{6\times 6} & \mathbf{K}_{11}^{(ke)} & \mathbf{K}_{12}^{(ke)} \\ \mathbf{K}_{21}^{(ie)} & \mathbf{K}_{21}^{(je)} & \mathbf{K}_{21}^{(ke)} & \mathbf{K}_{22}^{(ie)} + \mathbf{K}_{22}^{(je)} + \mathbf{K}_{22}^{(ke)} \end{bmatrix} \cdot \begin{bmatrix} \Delta \mathbf{t}_i \\ \Delta \mathbf{t}_j \\ \Delta \mathbf{t}_k \\ \Delta \mathbf{t}_e \end{bmatrix} \quad (11)$$

where $\Delta \mathbf{t}_i, \Delta \mathbf{t}_j, \Delta \mathbf{t}_k$ are the deflections at the leg clamping points, $\Delta \mathbf{t}_e$ is the deflection at the end-effector reference point, $\mathbf{W}_i, \mathbf{W}_j, \mathbf{W}_k$ are the wrenches at the leg clamping points, $\mathbf{W}_e$ is the total wrench applied to the end-effector from the virtual links side.

## 3. MSA MODELS OF MANIPULATOR JOINTS

### 3.1. Modeling of rigid joint

If adjacent links are connected by means of a rigid joint, the stiffness model must include relations describing two principal rules of structural mechanics: (a) displacement compatibility, (b) force equilibrium. If the nodes corresponding to the adjacent link denoted as $i, j, k, ...$, the first of these rules can be expressed as

$$\Delta \mathbf{t}_i = \Delta \mathbf{t}_j = \Delta \mathbf{t}_k = ... \quad (12)$$

that allows the user to merge corresponding columns in the global stiffness matrix at the assembling stage. The above equations can be also included in the stiffness model as an additional constraint without elimination redundant variables in the model. This constraint can be expressed as

$$\begin{bmatrix} \mathbf{I}_{6\times 6} & -\mathbf{I}_{6\times 6} \end{bmatrix}_{6\times 12} \cdot \begin{bmatrix} \Delta \mathbf{t}_i \\ \Delta \mathbf{t}_j \end{bmatrix} = \mathbf{0}_{6\times 1} \quad (13)$$

in the case of two adjacent links with the rigid connection of the nodes $i, j$, and as

$$\begin{bmatrix} \mathbf{I}_{6\times 6} & \mathbf{0}_{6\times 6} & -\mathbf{I}_{6\times 6} \\ \mathbf{0}_{6\times 6} & \mathbf{I}_{6\times 6} & -\mathbf{I}_{6\times 6} \end{bmatrix}_{12\times 18} \cdot \begin{bmatrix} \Delta \mathbf{t}_i \\ \Delta \mathbf{t}_j \\ \Delta \mathbf{t}_k \end{bmatrix} = \mathbf{0}_{12\times 1} \quad (14)$$

in the case of three adjacent links with the rigid connection of the nodes $i, j, k$.

The second rule (the force equilibrium) leads to

$$\mathbf{W}_i + \mathbf{W}_j + \mathbf{W}_k + ... = \mathbf{0} \quad (15)$$

that directly follows from the Newton's third law. In contrast to conventional approach, here this relation is included in the stiffness model as an additional constraint without elimination redundant variables. In the case of two adjacent links with rigid connection of the nodes $i, j$, the wrenches $\mathbf{W}_i, \mathbf{W}_j$ can be expressed in the matrix form as

$$\begin{bmatrix} \mathbf{I}_{6\times 6} & \mathbf{I}_{6\times 6} \end{bmatrix}_{6\times 12} \cdot \begin{bmatrix} \mathbf{W}_i \\ \mathbf{W}_j \end{bmatrix} = \mathbf{0}_{6\times 1} \quad (16)$$

Similarly, equation can be written for the case of three and more adjacent links with rigid connection. For example, for the case of three adjacent links it will have the form

$$\begin{bmatrix} \mathbf{I}_{6\times 6} & \mathbf{I}_{6\times 6} & \mathbf{I}_{6\times 6} \end{bmatrix}_{12\times 18} \cdot \begin{bmatrix} \mathbf{W}_i \\ \mathbf{W}_j \\ \mathbf{W}_k \end{bmatrix} = \mathbf{0}_{6\times 1} \quad (17)$$

Hence, the rigid joint imposes some additional constraints on the model variables $\Delta \mathbf{t}_i$ and $\mathbf{W}_i$. Number of these constraints is equal to 12 for the case of two links, 18 for three links, etc.

### 3.2. Modeling of passive joint

If adjacent links are connected by means of a passive joint, for two adjacent links $(ai)$, $(jb)$, the displacement compatibility condition of $\Delta \mathbf{t}_i, \Delta \mathbf{t}_j$ must be replaced by

$$\mathbf{\Lambda}_{ij}^r \cdot (\Delta \mathbf{t}_i - \Delta \mathbf{t}_j) = \mathbf{0} \quad (18)$$

where $\mathbf{\Lambda}_{ij}^r$ is $6 \times 6$ rank-deficient matrix that defines directions for the passive joint that do not admit free relative motions. Let us define the orthonormal basis $\mathbf{u}_1, \mathbf{u}_2, ..., \mathbf{u}_6$ associated with the passive joint in a such way that the unit vectors $\mathbf{u}_1, ..., \mathbf{u}_r$ describe the directions of the rigid connection and the unit vector $\mathbf{u}_{r+1}, ..., \mathbf{u}_6$ correspond to the passive connection allowing free relative motions of the links. This definition allows us to present the matrix $\mathbf{\Lambda}^r$ as

$$\mathbf{\Lambda}^r = \begin{bmatrix} \mathbf{u}_1, ..., \mathbf{u}_r, | \mathbf{0}_{6\times 1}, ..., \mathbf{0}_{6\times 1} \end{bmatrix}_{6\times 6}^T \quad (19)$$

that corresponds to the set of linear equations

$$\mathbf{u}_k^T \cdot (\Delta \mathbf{t}_i - \Delta \mathbf{t}_j) = 0, \quad \forall k = \overline{1, r} \quad (20)$$

Using the matrix $\mathbf{\Lambda}_{ij}^r$, the deflection compatibility constraint may be written in the form

$$\begin{bmatrix} \mathbf{\Lambda}_{ij}^r & -\mathbf{\Lambda}_{ij}^r \end{bmatrix}_{6\times 12} \cdot \begin{bmatrix} \Delta \mathbf{t}_i \\ \Delta \mathbf{t}_j \end{bmatrix} = \mathbf{0}_{6\times 1} \quad (21)$$

that may be further reduced to a full-rank matrix equation

$$\begin{bmatrix} \mathbf{\Lambda}_{*ij}^r & -\mathbf{\Lambda}_{*ij}^r \end{bmatrix}_{r\times 12} \cdot \begin{bmatrix} \Delta \mathbf{t}_i \\ \Delta \mathbf{t}_j \end{bmatrix} = \mathbf{0}_{r\times 1} \quad (22)$$

that contains a rectangular matrix $\mathbf{\Lambda}_{*ij}^r$ of the rank $r$. In the simple cases, the matrix $\mathbf{\Lambda}_{*ij}^r$ can be easily derived from $\mathbf{\Lambda}_{ij}^r$ by simple elimination of zero rows. In the general case, the reduced matrix $\mathbf{\Lambda}_{*ij}^r$ is formed as follows

$$\mathbf{\Lambda}_*^r = \begin{bmatrix} \mathbf{u}_1, ..., \mathbf{u}_r \end{bmatrix}_{r\times 6}^T \quad (23)$$

The force equilibrium condition is presented here as

$$\mathbf{\Lambda}_{ij}^r \cdot (\mathbf{W}_i + \mathbf{W}_j) = \mathbf{0} \quad (24)$$

that after elimination zero lines gives the second constraint

$$\begin{bmatrix} \mathbf{\Lambda}_{*ij}^r & \mathbf{\Lambda}_{*ij}^r \end{bmatrix}_{r\times 12} \cdot \begin{bmatrix} \mathbf{W}_i \\ \mathbf{W}_j \end{bmatrix} = \mathbf{0}_{r\times 1} \quad (25)$$

In addition, it is necessary to take into account that passive joints do not transmit the force/torque in the direction corresponding to zero columns in the matrix $\mathbf{\Lambda}_{ij}^r$, which yields the following equations

$$\mathbf{\Lambda}_{ij}^p \cdot \mathbf{W}_i = \mathbf{0}; \quad \mathbf{\Lambda}_{ij}^p \cdot \mathbf{W}_j = \mathbf{0} \quad (26)$$

Here, the matrix $\mathbf{\Lambda}_{ij}^p$ defines the passive joint axes and contains the remaining unit vectors of the basis $\mathbf{u}_1, \mathbf{u}_2, ..., \mathbf{u}_6$ that are not included in the $\mathbf{\Lambda}_{ij}^r$:

$$\Lambda^p = \begin{bmatrix} \mathbf{0}_{6\times 1}, ..., \mathbf{0}_{6\times 1}, | \mathbf{u}_{r+1}, ..., \mathbf{u}_6 \end{bmatrix}^T_{6\times 6} \quad (27)$$

Similar to above, equation (26) can be easily reduced by simple replacing of the singular $6 \times 6$ matrix $\Lambda^p_{ij}$ by the full-rank matrix $\Lambda^p_{*ij}$ of the size $p \times 6$, $p = 6 - r$ that contains non-zero column only

$$\Lambda^p_* = \begin{bmatrix} \mathbf{u}_{r+1}, ..., \mathbf{u}_6 \end{bmatrix}^T_{p\times 6} \quad (28)$$

The latter allows us to present the constraint as

$$\begin{bmatrix} \Lambda^p_{*ij} & \mathbf{0} \\ \mathbf{0} & \Lambda^p_{*ij} \end{bmatrix}_{2p\times 12} \cdot \begin{bmatrix} \mathbf{W}_i \\ \mathbf{W}_j \end{bmatrix} = \mathbf{0}_{2p\times 1} \quad (29)$$

It should be noted that all passive connections should be treated separately (in contrast to rigid connections).

### 3.3. Modeling of elastic joint

If the adjacent links are connected by means of an elastic joint that can be also treated as passive compliant joint with springs. In this case, the deflection compatibility condition (22) remains the same, it ensures equality for $r$ components of deflection vectors $\Delta \mathbf{t}_i$, $\Delta \mathbf{t}_j$. However, the force equilibrium condition must be slightly revised:

$$\mathbf{W}_i + \mathbf{W}_j = \mathbf{0};$$
$$\Lambda^e_{*ij} \cdot \mathbf{W}_i = \mathbf{K}^e_{ij} \cdot \Lambda^e_{*ij} \cdot (\Delta \mathbf{t}_i - \Delta \mathbf{t}_j) \quad (30)$$

where the matrix $\Lambda^e_{*ij}$ of size $e \times 6$, $e = 6 - r$ corresponds to the non-rigid directions of the joint (similar to $\Lambda^p_{*ij}$) and $\mathbf{K}^e_{ij}$ is $e \times e$ stiffness matrix describing elastic properties of the joint. So, the force equilibrium condition can be presented as the following matrix constraint

$$\begin{bmatrix} \mathbf{0}_{6\times 6} & \mathbf{0}_{6\times 6} & \mathbf{I}_{6\times 6} & \mathbf{I}_{6\times 6} \\ \mathbf{K}^e_{ij}\Lambda^e_{*ij} & -\mathbf{K}^e_{ij}\Lambda^e_{*ij} & \Lambda^e_{*ij} & \mathbf{0}_{e\times 6} \end{bmatrix} \begin{bmatrix} \Delta \mathbf{t}_i \\ \Delta \mathbf{t}_j \\ \mathbf{W}_i \\ \mathbf{W}_j \end{bmatrix} = \mathbf{0}_{(e+6)\times 1} \quad (31)$$

It should be stressed that the above expressions are valid for the so-called "no preloading case", when equal deflections $\Delta \mathbf{t}_i = \Delta \mathbf{t}_j$ does not generate the elastic forces, i.e. $\Lambda^e_{*ij} \cdot \mathbf{W}_i = \Lambda^e_{*ij} \cdot \mathbf{W}_j = \mathbf{0}$. In the case when the springs are initially preloaded by the wrench $\mathbf{W}^0_{ij}$, the force equilibrium equation must be presented in the form

$$\mathbf{W}_i + \mathbf{W}_j = \mathbf{0};$$
$$\Lambda^e_{*ij} \cdot (\mathbf{W}_i - \mathbf{W}^0_{ij}) = \mathbf{K}^e_{ij} \cdot \Lambda^e_{*ij} \cdot (\Delta \mathbf{t}_i - \Delta \mathbf{t}_j) \quad (32)$$

that leads to the following linear matrix constraint

$$\begin{bmatrix} \mathbf{0}_{6\times 6} & \mathbf{0}_{6\times 6} & \mathbf{I}_{6\times 6} & \mathbf{I}_{6\times 6} \\ \mathbf{K}^e_{ij}\Lambda^e_{*ij} & -\mathbf{K}^e_{ij}\Lambda^e_{*ij} & \Lambda^e_{*ij} & \mathbf{0}_{e\times 6} \end{bmatrix} \begin{bmatrix} \Delta \mathbf{t}_i \\ \Delta \mathbf{t}_j \\ \mathbf{W}_i \\ \mathbf{W}_j \end{bmatrix} = \begin{bmatrix} \mathbf{0}_{6\times 1} \\ \Lambda^e_{*ij}\mathbf{W}^0_{ij} \end{bmatrix} \quad (33)$$

Hence, the elastic joint may be included in the global stiffness model similarly to the passive one minor modifications of the corresponding matrix constraints.

### 3.4. Modeling of actuated joint

The actuated joint ensures transmission of the force/torque between the manipulator links. In the frame of MSA technique, it can be presented either as a rigid, passive or elastic connection. In the first two cases, the actuating effort is transmitting in the direction corresponding to the vectors $\mathbf{u}_1, ..., \mathbf{u}_r$ while in the last case transmission can be performed in the direction defined by the vectors $\mathbf{u}_{r+1}, ..., \mathbf{u}_6$. Hence, the actuated joint may be included in the global stiffness model as a set of two linear matrix constraints describing the deflection compatibility and force equilibrium, which slightly differ depending on the accepted joint idealization.

## 4. BOUNDARY CONDITIONS AND LOADINGS

### 4.1. Modeling of rigid support

The rigid connection of the link (*jb*) to the robot base can be presented as a special case of the rigid joint with eliminated link (*ai*) and zero deflection $\Delta \mathbf{t}_j = \mathbf{0}$. This simplifies the deflection compatibility constraint (13) down to

$$[\mathbf{I}_{6\times 6}] \cdot [\Delta \mathbf{t}_j] = \mathbf{0}_{6\times 1} \quad (34)$$

that contains 6 linear equations to be included in the global stiffness model. Corresponding reaction wrench at the rigid support may be computed as

$$\mathbf{W}_j = \mathbf{K}^{(jb)}_{12} \cdot \Delta \mathbf{t}_b \quad (35)$$

where the deflection $\Delta \mathbf{t}_b$ is obtained from solution of the global stiffness equations.

### 4.2. Modeling of passive support

The passive connection of the link (*jb*) to the robot base can be presented as a special case of the passive joint with eliminated link (*ai*) and zero deflection in the non-passive directions $\Lambda^r_{*ij} \cdot \Delta \mathbf{t}_j = \mathbf{0}$. This simplifies the deflection compatibility constraint (22) down to

$$[\Lambda^r_{*ij}]_{r\times 6} \cdot [\Delta \mathbf{t}_j] = \mathbf{0}_{r\times 1} \quad (36)$$

that contributes $r$ linear equations to the global stiffness model. In addition, it is necessary to take into account that the wrench components for the passive directions are equal to zero, which yields the following constraint

$$\Lambda^p_{ij} \cdot \mathbf{W}_j = \mathbf{0}_{p\times 1} \quad (37)$$

that contributes another $p$ equations to the global stiffness model. Totally, this gives $r + p = 6$ independent linear constraints. Corresponding reaction wrench at the passive support may be computed as

$$\mathbf{W}_j = \mathbf{K}^{(jb)}_{11} \cdot \Delta \mathbf{t}_j + \mathbf{K}^{(jb)}_{12} \cdot \Delta \mathbf{t}_b \quad (38)$$

where the deflections $\Delta \mathbf{t}_j$, $\Delta \mathbf{t}_b$ are obtained from solution of the global stiffness equations.

### 4.3. Modeling of elastic support

Similarly, the elastic connection of the link (*jb*) to the robot base can be presented as a special case of the elastic joint with eliminated link (*ai*) and zero deflection in the non-elastic directions $\Lambda^r_{*ij} \cdot \Delta \mathbf{t}_j = \mathbf{0}$. This allows us to use the same deflection compatibility constraint (36) as above, which contributes $r$ linear equations to the global stiffness model. In addition, it is necessary to take into account that the wrench components corresponding to the non-rigid directions are produced by the elastic forces satisfying the Hook's law,

$$\Lambda^e_{*ij} \cdot (\mathbf{W}_j - \mathbf{W}^0_{ij}) = \mathbf{K}^e_{ij} \cdot \Lambda^e_{*ij} \cdot \Delta \mathbf{t}_j \quad (39)$$

The latter leads to the following constraint equation

$$\begin{bmatrix} -\mathbf{K}^e_{ij} \cdot \mathbf{\Lambda}^e_{*ij} & \mathbf{\Lambda}^e_{*ij} \end{bmatrix}_{e \times 12} \cdot \begin{bmatrix} \Delta \mathbf{t}_j \\ \mathbf{W}_j \end{bmatrix} = \begin{bmatrix} \mathbf{\Lambda}^e_{*ij} \mathbf{W}^0_{ij} \end{bmatrix}_{e \times 1} \quad (40)$$

that contributes another $e$ equations to the global stiffness model. Totally, this gives $r + e = 6$ independent linear constraints. To compute the reaction wrench at the elastic support the same expression (38) can be used.

### 4.4. Including the external loading

In robotic manipulators, both serial and parallel, there is at least one node that is not connected directly to the robot base. It corresponds to the end-effector that interacts with robot environment by applying the force/torque to the external objects. For the global stiffness model, the end-effector produces the boundary conditions that should include the vector of external wrench $\mathbf{w}$, which is also necessary for computation of the Cartesian stiffness matrix.

To take into account the external loading $\mathbf{w}$, the global stiffness model must be completed by the linear constraint derived from the force equilibrium at the node $e$, i.e. $\mathbf{W}_i + \mathbf{W}_j = \mathbf{W}_e$, which can be rewritten in the form

$$\begin{bmatrix} \mathbf{I}_{6 \times 6} & \mathbf{I}_{6 \times 6} \end{bmatrix}_{6 \times 12} \cdot \begin{bmatrix} \mathbf{W}_i \\ \mathbf{W}_j \end{bmatrix}_{12 \times 1} = \begin{bmatrix} \mathbf{W}_e \end{bmatrix}_{6 \times 1} \quad (41)$$

Similarly, it is possible to derive the constraints for the case of three and more adjacent links, for instance

$$\begin{bmatrix} \mathbf{I}_{6 \times 6} & \mathbf{I}_{6 \times 6} & \mathbf{I}_{6 \times 6} \end{bmatrix}_{6 \times 18} \cdot \begin{bmatrix} \mathbf{W}_i \\ \mathbf{W}_j \\ \mathbf{W}_k \end{bmatrix}_{18 \times 1} = \begin{bmatrix} \mathbf{W}_e \end{bmatrix}_{6 \times 1} \quad (42)$$

This technique is applied to all nodes with external wrenches.

## 5. AGGREGATION OF MSA MODEL COMPONENTS

In contrast to large mechanical structures consisting of a huge number of flexible components, the robotic manipulator is rather simple for MSA analysis. Since the number of flexible links in manipulator is relatively small, the assembling stage can be simplified and the columns/rows merging operations can be avoided and replaced by adding relevant constraints.

As it was shown above, the MSA equations for robotic manipulator are derived from three main sources: (i) link models, (ii) joint models and (iii) boundary conditions. The first of them describes the force-displacement relations for all links (both flexible and rigid) yielding 12 scalar equations per link relating 24 variables $\{\Delta \mathbf{t}_i, \Delta \mathbf{t}_j, \mathbf{W}_i, \mathbf{W}_j\}$. The second group of equations ensures the displacement compatibility and force/torque equilibrium for each internal connection (both rigid, passive and elastic), it includes two types of relations written for $\{\Delta \mathbf{t}_i, \Delta \mathbf{t}_j, ...\}$ and $\{\mathbf{W}_i, \mathbf{W}_j, ...\}$ separately. Independent of the connection type, each joint provides 12 scalar equations if it connects two links, 18 scalar equations for the connection of three links, etc. The third group of equations is issued from the manipulator connections to the environment, they are presented as the force/displacement constraints for certain nodes that give up to 6 scalar equations for $\{\Delta \mathbf{t}_i, \Delta \mathbf{t}_j, ...\}$ or at least 6 scalar equations for $\{\mathbf{W}_i, \mathbf{W}_j, ...\}$ depending on the connection type.

To derive the global stiffness model of the considered manipulator, let us combine all displacements and all wrenches in single vectors and denote them as $\{\Delta \mathbf{t}_i\}$ and $\{\mathbf{W}_i\}$. This allows us to present the desired model in the form of the following block-matrix equation of sparse structure

$$\begin{bmatrix} \{\mathbf{K}^{ij}_{12 \times 12}\} & \{-\mathbf{I}_{12 \times 12}\} \\ \{\mathbf{D}^{ij}_{6 \times 6}, -\mathbf{I}_{6 \times 6}\} & \\ & \{\mathbf{I}_{6 \times 6}, \mathbf{D}^{ij\,T}_{6 \times 6}\} \\ \{\mathbf{I}_{6 \times 6}, -\mathbf{I}_{6 \times 6}\} & \\ & \{\mathbf{I}_{6 \times 6}, \mathbf{I}_{6 \times 6}\} \\ \{\mathbf{\Lambda}^r_{*ij}, -\mathbf{\Lambda}^r_{*ij}\} & \\ & \{\mathbf{\Lambda}^r_{*ij}, \mathbf{\Lambda}^r_{*ij}\} \\ & \{\mathbf{\Lambda}^P_{*ij}\} \\ \{\mathbf{\Lambda}^r_{*ij}, -\mathbf{\Lambda}^r_{*ij}\} & \\ & \{\mathbf{I}_{6 \times 6}, \mathbf{I}_{6 \times 6}\} \\ \{\mathbf{K}^e_{ij}\mathbf{\Lambda}^e_{*ij}, -\mathbf{K}^e_{ij}\mathbf{\Lambda}^e_{*ij}\} & \mathbf{\Lambda}^e_{*ij} \\ \{\mathbf{I}_{6 \times 6}\} & \\ \{\mathbf{\Lambda}^r_{*ij}\} & \\ & \{\mathbf{\Lambda}^P_{*ij}\} \\ \{\mathbf{\Lambda}^r_{*ij}\} & \\ \{\mathbf{K}^e_{ij}\mathbf{\Lambda}^e_{*ij}\} & \mathbf{\Lambda}^e_{*ij} \\ & \{-\mathbf{I}_{6 \times 6}\} \end{bmatrix} \cdot \begin{bmatrix} \{\Delta \mathbf{t}_i\} \\ \{\mathbf{W}_i\} \end{bmatrix} = \begin{bmatrix} 0 \\ 0 \\ 0 \\ 0 \\ \{\mathbf{W}^{ext}_i\} \\ 0 \\ \{\mathbf{W}^{ext}_{*i}\} \\ 0 \\ 0 \\ \{\mathbf{W}^{ext}_i\} \\ \{\mathbf{W}^0_{ij}\} \\ 0 \\ 0 \\ 0 \\ 0 \\ \{\mathbf{W}^0_{ij}\} \\ \{\mathbf{W}^{ext}_i\} \end{bmatrix} \quad (43)$$

After rearranging the matrix rows and introducing relevant definitions for the blocks, the global stiffness model can be presented a

$$\begin{bmatrix} \mathbf{S}_{agr} & \mathbf{K}_{agr} \\ \mathbf{E}_{agr} & \mathbf{F}_{agr} \end{bmatrix} \cdot \begin{bmatrix} \mathbf{W}_{agr} \\ \Delta \mathbf{t}_{agr} \end{bmatrix} = \begin{bmatrix} \mathbf{b}_{agr} \\ \mathbf{W}_{ext} \end{bmatrix} \quad (44)$$

where the matrices $\mathbf{S}_{agr}$, $\mathbf{K}_{agr}$, $\mathbf{E}_{agr}$, $\mathbf{F}_{agr}$, $\mathbf{b}_{agr}$, $\mathbf{W}_{ext}$ are generated using relevant models, while the vectors $\mathbf{W}_{agr}$ and $\Delta \mathbf{t}_{agr}$ contain all variables describing the wrenches and displacements, respectively. To find the desired Cartesian stiffness matrix, let us divide the node displacement variables $\Delta \mathbf{t}_{agr}$ into two groups $\Delta \mathbf{t}_m$ and $\Delta \mathbf{t}_e$ corresponding to the manipulator internal nodes and the end-effector node, where the external wrench $\mathbf{W}_e$ is applied. The latter allows us to rewrite the system in the form

$$\begin{bmatrix} \mathbf{S}_{agr} & \mathbf{K}_m & \mathbf{K}_e \\ \mathbf{E}_m & \mathbf{F}_m & \mathbf{C}_e \\ \mathbf{E}_e & \mathbf{F}_e & \mathbf{D} \end{bmatrix} \cdot \begin{bmatrix} \mathbf{W}_{agr} \\ \Delta \mathbf{t}_m \\ \Delta \mathbf{t}_e \end{bmatrix} = \begin{bmatrix} \mathbf{b}_{agr} \\ \mathbf{W}_m \\ \mathbf{W}_e \end{bmatrix} \quad (45)$$

and further, present it as

$$\begin{bmatrix} \mathbf{A} & \mathbf{B} \\ \mathbf{C} & \mathbf{D} \end{bmatrix} \cdot \begin{bmatrix} \mathbf{\mu} \\ \Delta \mathbf{t}_e \end{bmatrix} = \begin{bmatrix} \mathbf{b} \\ \mathbf{W}_e \end{bmatrix} \quad (46)$$

where all internal variables are included in the vector $\mathbf{\mu} = col(\mathbf{W}_{agr}, \Delta \mathbf{t}_m)$ an

$$\mathbf{A} = \begin{bmatrix} \mathbf{S}_{agr} & \mathbf{K}_m \\ \mathbf{E}_m & \mathbf{F}_m \end{bmatrix}; \quad \mathbf{B} = \begin{bmatrix} \mathbf{K}_e \\ \mathbf{C}_e \end{bmatrix}; \quad \mathbf{C} = \begin{bmatrix} \mathbf{E}_e & \mathbf{F}_e \end{bmatrix}; \quad \mathbf{b} = \begin{bmatrix} \mathbf{b}_{agr} \\ \mathbf{W}_m \end{bmatrix} \quad (47)$$

Using the obtained system, the desired Cartesian stiffness matrix can be computed as

$$\mathbf{K}_C = \mathbf{D} - \mathbf{C} \cdot \mathbf{A}^{-1} \cdot \mathbf{B} \quad (48)$$

It is worth mentioning that $\mathbf{A}^{-1}$ usually exists (if the manipulator does not include redundant passive joints), otherwise, the pseudo-inverse should be used that distribute deflections between the redundant joints.

# 6. ILLUSTRATIVE EXAMPLES: MSA-BASED ANALYSIS OF NAVARO MANIPULATOR

To demonstrate the utility of the MSA-based technique for robotics, let us apply it to the stiffness analysis of NaVaRo robot (Figure 1). It is composed of three identical legs and a moving platform formed of three segments rigidly linked at the central point. Each leg consists of four non-rigid links connected by five revolute joints to create a parallelogram linkage. Among them, there are four passive joints and one actuated joint connected to the motor.

To apply the MSA technique, let us split the manipulator into four parts: three kinematically identical legs and a mobile platform. For convenience, let us divide the longest link of the leg into two rigidly connected parts. This allows us to present each leg (Figure 2) as a set of five flexible links (1,2), (3,4), (5,6), (7,8), (9,e) with passive connections of the nodes <2,3>, <4,5>, <6,7> and rigid connection of the node <6,9>. Besides, the boundary conditions induced by the motor can be specified as the passive connection of the nodes <0,8> and elastic connection of the nodes <0,1>.

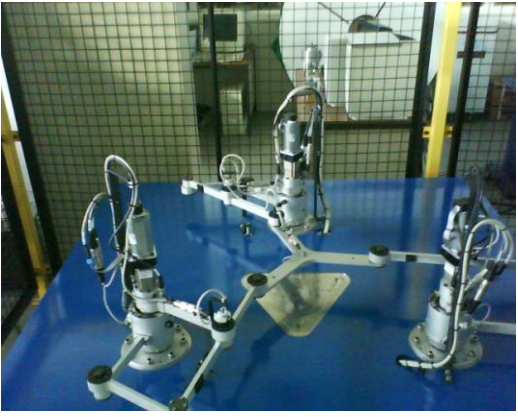

Figure 1. Parallel planar manipulator NaVaRo.

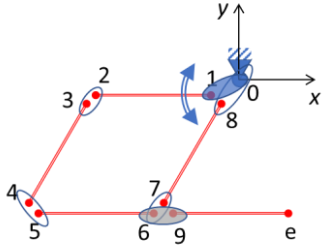

Figure 2. MSA-based representation of the NaVaRo leg.

In the frame of the adopted notation, each link can be described by the stiffness model

$$\begin{bmatrix} \mathbf{W}_i \\ \mathbf{W}_j \end{bmatrix} = \begin{bmatrix} \mathbf{K}_{11}^{(ij)} & \mathbf{K}_{12}^{(ij)} \\ \mathbf{K}_{21}^{(ij)} & \mathbf{K}_{22}^{(ij)} \end{bmatrix}_{12\times 12} \cdot \begin{bmatrix} \Delta \mathbf{t}_i \\ \Delta \mathbf{t}_j \end{bmatrix} \quad (49)$$

$(i,j) \in \{(1,2),(3,4),(5,6),(7,8),(9,e)\}$

where all matrices are presented in the global coordinate system. After assembling (49) one can get $\mathbf{K}_{links}$ aggregated stiffness matrix of a size $60 \times 60$ allowing to present the manipulator leg stiffness model in the following form

$$\begin{bmatrix} -\mathbf{I}_{60\times 60} & \mathbf{K}_{links} \end{bmatrix}_{60\times 120} \cdot \begin{bmatrix} \mathbf{W}_{agr} \\ \Delta \mathbf{t}_{agr} \end{bmatrix}_{120\times 1} = \mathbf{0}_{60\times 1} \quad (50)$$

where $\Delta \mathbf{t}_{agr}$ and $\mathbf{W}_{agr}$ are the aggregated displacement and wrenches defined as

$$\Delta \mathbf{t}_{agr} = col(\Delta \mathbf{t}_1, ..., \Delta \mathbf{t}_9, \Delta \mathbf{t}_e); \quad \mathbf{W}_{agr} = col(\mathbf{W}_1, ..., \mathbf{W}_9, \mathbf{W}_e) \quad (51)$$

Further, the stiffness model of the leg should be complemented by the second component that includes equation issued from the inter-link connections. For the considered manipulator, the links are connected by means of the rigid joint <6,9>, passive joints <2,3>, <4,5>, <6,7>, and elastic revolute joint <0,1>. The passive and elastic joints provide rotation around the z-axis, which corresponds to the constraint matrices of the following form

$$\mathbf{\Lambda}_*^r = \begin{bmatrix} 1 & 0 & 0 & 0 & 0 & 0 \\ 0 & 1 & 0 & 0 & 0 & 0 \\ 0 & 0 & 1 & 0 & 0 & 0 \\ 0 & 0 & 0 & 1 & 0 & 0 \\ 0 & 0 & 0 & 0 & 1 & 0 \end{bmatrix}; \quad \begin{array}{l} \mathbf{\Lambda}_*^p = [0\ 0\ 0\ 0\ 0\ 1] \\ \mathbf{\Lambda}_*^e = [0\ 0\ 0\ 0\ 0\ 1] \end{array} \quad (52)$$

Using this notation, the constraints imposed by the passive joints <2,3>, <4,5> can be presented as

$$\begin{bmatrix} \mathbf{\Lambda}_*^r & -\mathbf{\Lambda}_*^r \end{bmatrix} \cdot \begin{bmatrix} \Delta \mathbf{t}_i \\ \Delta \mathbf{t}_j \end{bmatrix} = \mathbf{0}_{5\times 1}; \quad \begin{bmatrix} \mathbf{\Lambda}_*^r & \mathbf{\Lambda}_*^r \end{bmatrix} \cdot \begin{bmatrix} \mathbf{W}_i \\ \mathbf{W}_j \end{bmatrix} = \mathbf{0}_{5\times 1}$$

$$\begin{bmatrix} \mathbf{\Lambda}^p & \mathbf{0}_{1\times 6} \\ \mathbf{0}_{1\times 6} & \mathbf{\Lambda}^p \end{bmatrix} \cdot \begin{bmatrix} \mathbf{W}_i \\ \mathbf{W}_j \end{bmatrix} = \mathbf{0}_{2\times 1}; \quad (i,j) \in \{(2,3),(4,5)\} \quad (53)$$

The rigid joint constraint <6,9>, which ensures stiff connection of the links (5,6) and (9,e), should be combined with the passive joint constraint <6,7> that yields the following equations

$$\begin{bmatrix} \mathbf{\Lambda}_*^r & -\mathbf{\Lambda}_*^r \end{bmatrix} \cdot \begin{bmatrix} \Delta \mathbf{t}_6 \\ \Delta \mathbf{t}_7 \end{bmatrix} = \mathbf{0}_{5\times 1}; \quad \begin{bmatrix} \mathbf{I}_{6\times 6} & -\mathbf{I}_{6\times 6} \end{bmatrix} \cdot \begin{bmatrix} \Delta \mathbf{t}_6 \\ \Delta \mathbf{t}_9 \end{bmatrix} = \mathbf{0}_{6\times 1}$$

$$\begin{bmatrix} \mathbf{\Lambda}_*^r & \mathbf{\Lambda}_*^r & \mathbf{\Lambda}_*^r \end{bmatrix} \cdot \begin{bmatrix} \mathbf{W}_6 \\ \mathbf{W}_7 \\ \mathbf{W}_9 \end{bmatrix} = \mathbf{0}_{5\times 1}; \quad \begin{bmatrix} \mathbf{\Lambda}_*^p & \mathbf{\Lambda}_*^p \end{bmatrix} \cdot \begin{bmatrix} \mathbf{W}_6 \\ \mathbf{W}_9 \end{bmatrix} = \mathbf{0}_{1\times 1}; \quad (54)$$

$$\mathbf{\Lambda}^p \cdot \mathbf{W}_7 = \mathbf{0}_{1\times 1}$$

The third model component includes the boundary conditions. For the adopted actuation mode when the motor is connected to the link (1,2) via an elastic transmission and the link (7,8) is passively connected to the base, the boundary condition at the node <0,1,8> is presented in the form

$$\mathbf{\Lambda}_*^r \Delta \mathbf{t}_1 = \mathbf{0}_{5\times 1}; \quad \mathbf{\Lambda}_*^r \Delta \mathbf{t}_8 = \mathbf{0}_{5\times 1}$$
$$\mathbf{\Lambda}^p \cdot \mathbf{W}_8 = \mathbf{0}_{1\times 1}; \quad K_e \cdot \mathbf{\Lambda}_*^e \cdot \Delta \mathbf{t}_1 - \mathbf{\Lambda}_*^e \cdot \mathbf{W}_1 = 0 \quad (55)$$

where $\Delta \mathbf{t}_0$ is assumed to be equal to zero. The external loading can be introduced into the system as

$$\mathbf{W}_e = \mathbf{W}_{ext} \quad (56)$$

After assembling all equations describing the links elasticities, constraints and boundary conditions the aggregated stiffness model of the manipulator's leg is presented as follows

$$\begin{bmatrix} -\mathbf{I}_{60\times 60} & \mathbf{K}_{links} \\ \mathbf{0}_{31\times 60} & \mathbf{A}_{agr} \\ \mathbf{B}_{agr} & \mathbf{0}_{22\times 60} \\ \mathbf{C}_{agr} & \mathbf{D}_{agr} \\ \mathbf{E}_{agr} & \mathbf{0}_{6\times 60} \end{bmatrix}_{120\times 120} \cdot \begin{bmatrix} \mathbf{W}_{agr} \\ \Delta \mathbf{t}_{agr} \end{bmatrix}_{120\times 1} = \begin{bmatrix} \mathbf{0}_{114\times 1} \\ \mathbf{W}_e \end{bmatrix}_{120\times 1} \quad (57)$$

where

$$\mathbf{A}_{agr} = \begin{bmatrix} 0 & \Lambda_*^r & -\Lambda_*^r & 0 & 0 & 0 & 0 & 0 & 0 & 0 \\ 0 & 0 & 0 & \Lambda_*^r & -\Lambda_*^r & 0 & 0 & 0 & 0 & 0 \\ 0 & 0 & 0 & 0 & 0 & \Lambda_*^r & -\Lambda_*^r & 0 & 0 & 0 \\ 0 & 0 & 0 & 0 & 0 & 0 & \mathbf{I} & 0 & -\mathbf{I} & 0 \\ 0 & 0 & 0 & 0 & 0 & 0 & 0 & \Lambda_*^r & 0 & 0 \\ \Lambda_*^r & 0 & 0 & 0 & 0 & 0 & 0 & 0 & 0 & 0 \end{bmatrix}_{31 \times 60} \quad (58)$$

$$\mathbf{B}_{agr} = \begin{bmatrix} 0 & \Lambda_*^r & \Lambda_*^r & 0 & 0 & 0 & 0 & 0 & 0 & 0 \\ 0 & \Lambda_*^p & 0 & 0 & 0 & 0 & 0 & 0 & 0 & 0 \\ 0 & 0 & \Lambda_*^p & 0 & 0 & 0 & 0 & 0 & 0 & 0 \\ 0 & 0 & 0 & \Lambda_*^r & \Lambda_*^r & 0 & 0 & 0 & 0 & 0 \\ 0 & 0 & 0 & \Lambda_*^p & 0 & 0 & 0 & 0 & 0 & 0 \\ 0 & 0 & 0 & 0 & \Lambda_*^p & 0 & 0 & 0 & 0 & 0 \\ 0 & 0 & 0 & 0 & 0 & \Lambda_*^r & \Lambda_*^r & 0 & \Lambda_*^r & 0 \\ 0 & 0 & 0 & 0 & 0 & \Lambda_*^p & 0 & 0 & \Lambda_*^p & 0 \\ 0 & 0 & 0 & 0 & 0 & 0 & \Lambda_*^p & 0 & 0 & 0 \end{bmatrix}_{22 \times 60} \quad (59)$$

$$\mathbf{C}_{agr} = \begin{bmatrix} K_e \Lambda_*^e & 0 & 0 & 0 & 0 & 0 & 0 & 0 & 0 & 0 \end{bmatrix}_{1 \times 60} \quad (60)$$

$$\mathbf{D}_{agr} = \begin{bmatrix} -\Lambda_*^e & 0 & 0 & 0 & 0 & 0 & 0 & 0 & 0 & 0 \end{bmatrix}_{1 \times 60} \quad (61)$$

$$\mathbf{E}_{agr} = \begin{bmatrix} 0 & 0 & 0 & 0 & 0 & 0 & 0 & 0 & \mathbf{I}_{6 \times 6} \end{bmatrix}_{6 \times 60} \quad (62)$$

$$\mathbf{F}_{agr} = \begin{bmatrix} 0 & 0 & 0 & 0 & 0 & 0 & 0 & 0 & 0 & 0 \end{bmatrix}_{6 \times 60} \quad (63)$$

For computational convenience, this system can be also presented in the form (44), where

$$\mathbf{K}_{agr} = \begin{bmatrix} \mathbf{K}_{links} \\ \mathbf{A}_{agr} \\ \mathbf{0}_{22 \times 60} \\ \mathbf{D}_{agr} \end{bmatrix}_{114 \times 60} ; \quad \mathbf{S}_{agr} = \begin{bmatrix} -\mathbf{I}_{60 \times 60} \\ \mathbf{0}_{31 \times 60} \\ \mathbf{B}_{agr} \\ \mathbf{C}_{agr} \end{bmatrix}_{114 \times 60} \quad (64)$$

Further, after separating the node variables into two groups corresponding to the internal nodes $\Delta \mathbf{t}_m$ and to the end effector node $\Delta \mathbf{t}_e$ one can compute the desired stiffness matrix. After computing the matrices $\mathbf{K}_C$ for all three legs, they should be aggregated in a similar way, using obtained stiffness models of each leg as well as constraints (8), (10) imposed by the rigid platform with three passive connections. The latter are included in the model similarly to (53).

## 7. CONCLUSIONS

This paper deals with enhancement of the Matrix structural analysis for robotics applications, especially for the cases of complex and hybrid structures. In contrast to previous results, it is suitable for mixed architectures containing closed-loops, flexible links, rigid connections, passive and elastic joints with external loadings and preloadings. The paper describes in detail required matrix transformations that allow the user to obtain desired force-deflection relation and torque/deflections in internal mechanical elements. Its allows user straightforward aggregation of stiffness model equations avoiding traditional column/row merging procedures in the extended stiffness matrix. The particularities of this method are illustrated by an application example that deals with the stiffness analysis of NAVARO parallel manipulator.

## ACKNOWLEDGMENTS

The work presented in this paper was supported by the grant of Russian Science Foundation №17-19-01740.

## REFERENCES

Alici G, Shirinzadeh B. Enhanced stiffness modeling, identification and characterization for robot manipulators. Robotics, IEEE Transactions on. 2005;21:554-64.
Azulay H, Mahmoodi M, Zhao R, Mills JK, Benhabib B. Comparative analysis of a new 3×PPRS parallel kinematic mechanism. Robotics and Computer-Integrated Manufacturing. 2014;30:369-78.
Cammarata A. Unified formulation for the stiffness analysis of spatial mechanisms. Mechanism and Machine Theory. 2016;105:272-84.
Deblaise D, Hernot X, Maurine P. A systematic analytical method for PKM stiffness matrix calculation. ICRA 2006. p. 4213-9.
Detert T, Corves B. Extended Procedure for Stiffness Modeling Based on the Matrix Structure Analysis. Proceedings of The International Conference Robotics '16. Cham: Springer; 2017. p. 299-310.
Gonçalves RS, Carbone G, Carvalho JCM, Ceccarelli M. A comparison of stiffness analysis methods for robotic systems. International journal of mechanics and control. 2016;17:35-58.
Gosselin C. Stiffness mapping for parallel manipulators. Robotics and Automation, IEEE Transactions on. 1990;6:377-82.
Gosselin C, Zhang D. Stiffness analysis of parallel mechanisms using a lumped model. Int. J. of Robotics and Automation. 2002;17:17-27.
Klimchik A, Ambiehl A, Garnier S, Furet B, Pashkevich A. Efficiency evaluation of robots in machining applications using industrial performance measure. Robotics and Comp-Int. Man. 2017;48:12-29.
Klimchik A, Chablat D, Pashkevich A. Stiffness modeling for perfect and non-perfect parallel manipulators under internal and external loadings. Mechanism and Machine Theory. 2014;79:1-28.
Klimchik A, Furet B, Caro S, Pashkevich A. Identification of the manipulator stiffness model parameters in industrial environment. Mechanism and Machine Theory. 2015;90:1-22.
Klimchik A, Pashkevich A, Chablat D. CAD-based approach for identification of elasto-static parameters of robotic manipulators. Finite Elements in Analysis and Design. 2013;75:19-30.
Koseki Y, Tanikawa T, Koyachi N, Arai T. Kinematic analysis of a translational 3-dof micro-parallel mechanism using the matrix method. Advanced Robotics. 2002;16:251-64.
Li M, Wu H, Handroos H. Static stiffness modeling of a novel hybrid redundant robot machine. Fusion Engineering and Design. 2011;86:1838-42.
Liu H, Huang T, Chetwynd DG, Kecskeméthy A. Stiffness Modeling of Parallel Mechanisms at Limb and Joint/Link Levels. IEEE Transactions on Robotics. 2017;33:734-41.
Marie S, Courteille E, Maurine P. Elasto-geometrical modeling and calibration of robot manipulators: Application to machining and forming applications. Mech. and Machine Theory. 2013;69:13-43.
Pashkevich A, Klimchik A, Chablat D. Enhanced stiffness modeling of manipulators with passive joints. Mech. & Mach. Th.. 2011;46:662-79.
Pinto C, Corral J, Altuzarra O, Hernández A. A methodology for static stiffness mapping in lower mobility parallel manipulators with decoupled motions. Robotica. 2009;28:719-35.
Przemieniecki JS. Theory of matrix structural analysis: Courier Dover Publications; 1985.
Quennouelle C, Gosselin Cá. Stiffness matrix of compliant parallel mechanisms. Advances in Robot Kinematics: Analysis and Design: Springer; 2008. p. 331-41.
Salisbury JK. Active stiffness control of a manipulator in Cartesian coordinates. Decision and Control including the Symposium on Adaptive Processes, 1980 19th IEEE Conference on; 1980. p. 95-100.
Shi S, Wu H, Song Y, Handroos H, Li M, Cheng Y, et al. Static stiffness modelling of EAST articulated maintenance arm using matrix structural analysis method. Fusion Engineering and Design. 2017.
Slavkovic NR, Milutinovic DS, Glavonjic MM. A method for off-line compensation of cutting force-induced errors in robotic machining by tool path modification. The International Journal of Advanced Manufacturing Technology. 2013;70:2083-96.
Taghvaeipour A, Angeles J, Lessard L. On the elastostatic analysis of mechanical systems. Mech. and Machine Theory. 2012;58:202-16.
Wang YY, Huang T, Zhao XM, Mei JP, Chetwynd DG, Hu SJ. Finite Element Analysis and Comparison of Two Hybrid Robots-the Tricept and the TriVariant. IROS 2006. p. 490-5.
Yan SJ, Ong SK, Nee AYC. Stiffness analysis of parallelogram-type parallel manipulators using a strain energy method. Robotics and Computer-Integrated Manufacturing. 2016;37:13-22.
Yeo SH, Yang G, Lim WB. Design and analysis of cable-driven manipulators with variable stiffness. Mech.&Mach.Th.. 2013;69:230-44.